\documentclass[nojss]{jss}

\usepackage{thumbpdf,lmodern}

\usepackage{framed}
\usepackage{amsmath}
\usepackage{amssymb}
\usepackage{caption}

\usepackage[ruled]{algorithm2e}

\newcommand{\fct}[1]{\code{#1()}}

\author{Barinder Thind\\Simon Fraser University
   \And Sidi Wu\\Simon Fraser University
   \AND Richard Groenewald\\Columbia University
   \And Jiguo Cao\\Simon Fraser University} 
\Plainauthor{Barinder Thind, Sidi Wu, Richard Groenewald, Jiguo Cao}

\title{\pkg{FuncNN}: An \proglang{R} Package to Fit Deep Neural\\ Networks Using Generalized Input Spaces}
\Plaintitle{FuncNN: An R Package to Fit Deep Neural Networks Using Generalized Input Spaces}
\Shorttitle{The \pkg{FuncNN} Package}

\Abstract{
  Neural networks have excelled at regression and classification problems when the input space consists of scalar variables. As a result of this proficiency, several popular packages have been developed that allow users to easily fit these kinds of models. However, the methodology has excluded the use of functional covariates and to date, there exists no software that allows users to build deep learning models with this generalized input space. To the best of our knowledge, the functional neural network (\pkg{FuncNN}) library is the first such package in any programming language; the library has been developed for \proglang{R} and is built on top of the \pkg{keras} architecture. Throughout this paper, several functions are introduced that provide users an avenue to easily build models, generate predictions, and run cross-validations. A summary of the underlying methodology is also presented. The ultimate contribution is a package that provides a set of general modelling and diagnostic tools for data problems in which there exist both functional and scalar covariates.
}

\Keywords{functional inputs, deep learning, \proglang{R}, \pkg{keras}}
\Plainkeywords{functional inputs, deep learning, R, Keras}

\Address{
  Jiguo Cao\\
  Department of Statistics\\
  Faculty of Science\\
  Simon Fraser University\\
  8888 University Drive\\
  E-mail: \email{jiguo\_cao@sfu.ca}\\
  URL: \url{http://people.stat.sfu.ca/~cao/}
}

\begin{document}

%% -- Introduction -------------------------------------------------------------

%% - In principle "as usual".
%% - But should typically have some discussion of both _software_ and _methods_.
%% - Use \proglang{}, \pkg{}, and \code{} markup throughout the manuscript.
%% - If such markup is in (sub)section titles, a plain text version has to be
%%   added as well.
%% - All software mentioned should be properly \cite-d.
%% - All abbreviations should be introduced.
%% - Unless the expansions of abbreviations are proper names (like "Journal
%%   of Statistical Software" above) they should be in sentence case (like
%%   "generalized linear models" below).

\section[Introduction]{Introduction} \label{sec:intro}

In recent years, deep learning methodologies have become the standard approach for prediction problems. For example, \cite{he2016deep} broke previous benchmarks by developing deep neural networks with over a hundred hidden layers; this was made possible by an innovation brought upon by a simple adjustment of the algorithm -- adding the output of a previous layer back into the layers following it. Another set of examples can be found in any one of the annual ImageNet Large-Scale Visual Recognition Challenges \citep{krizhevsky2012imagenet, russakovsky2015imagenet} where neural networks are consistently on the podium. These innovations naturally sparked the development of software that supplied users with an arsenal of tools to apply these methods.

There is a wide array of software available in multiple programming languages that implements a variety of methods in deep learning and neural networks.  For example, \proglang{R} packages \pkg{nnet} \citep{nnet} and \pkg{neuralnet} \citep{neuralnet} are standard sets of tools in the training of and prediction from neural networks using the Broyden-Fletcher-Goldfarb-Shanno (BFGS) algorithm and backpropagation, respectively.  Both of these allow for some flexibility in terms of user selected activation functions and error measurements, but vastly more broad selections of tools are available within the \pkg{RSNNS} \citep{RSNNS} and \pkg{keras} \citep{keras} application programming interfaces (API), which may be used in both \proglang{R} and \proglang{Python}.  Both of these make detailed user customization possible and are intended to facilitate the development of deep learning models and methodologies.

Central to this paper is functional data analysis (FDA), an area of modern applied statistics that has witnessed swift development due to its ability to deal with a large and distinct class of statistical problems.  In particular, it is often the case that data is collected at high resolution at many points in space or time, resembling and being treated as a smooth function - this is known as functional data.  FDA is used instead of the standard approach of analysing single points when we are interested in framing an analysis of a set of points as if they are samples from a specific curve.

Presently, the primary software for FDA in any programming language is the \pkg{fda} package \citep{fdapkg} in \proglang{R}, designed to accompany the textbooks of \cite{Ramsay05} and \cite{ramsay2009}.  The main functionality within the package revolves around functional regression, allowing users to fit regression models where one or several of the response and explanatory variables are functions instead of scalar valued.  Several packages in \proglang{R} build upon this base by extending standard statistical methods to functional data.  For example, \pkg{funHDDC} \citep{funHDDC} and \pkg{funFEM} \citep{funFEM} handle functional clustering, \pkg{FDboost} \citep{FDboost} applies gradient boosting in functional regression and \pkg{GPFDA} \citep{GPFDA} incorporates functional regression as the mean response in Gaussian processes, among other packages. Another library, \pkg{fda.usc} \citep{fda.usc}, streamlines the fitting process of many common functional modelling approaches such as non-parametric functional regression \citep{ferraty2006nonparametric} and functional partial least squares methods \citep{aneiros2006semi}. Additionally, the \pkg{BFDA} package \citep{BFDA} in \proglang{MATLAB} provides a library for Bayesian FDA.

To the best of our knowledge, the applications of these packages do not extend to include deep learning models.  For this reason, this paper introduces the \pkg{FuncNN} package, which allows users to include functions as input variables in neural networks and implements the theoretical discussion in \cite{thind2020deep} and \cite{rossi2005functional}. This methodology has been shown to outperform many other functional and multivariate methods in a number of real world and simulated examples as seen in \cite{thind2020deep}. 

The \pkg{FuncNN} package introduces several functions. The primary function \fct{fnn.fit}, grants users the ability to effortlessly generate models for their data; they also gain access to several customization options in the form of hyperparameters. The output of this function can be acted upon by several other functions within the package such as \fct{fnn.predict} and \fct{fnn.fnc} which serve to be predictive and visualization functions. Moreover, we introduce tuning and cross-validation functions aptly named \fct{fnn.tune} and \fct{fnn.cv}, respectively. The package is readily available on CRAN can be installed as follows:
\begin{CodeChunk}
\begin{CodeInput}
R> install.packages("FuncNN")
\end{CodeInput}
\end{CodeChunk}
Or through GitHub using the \pkg{devtools} \citep{devtools} package:
\begin{CodeChunk}
\begin{CodeInput}
R> devtools::install_github("b-thi/FuncNN")
\end{CodeInput}
\end{CodeChunk}
The rest of the article is organized as follows: \hyperref[sec:bg]{Section 2} explores some of the theoretical background associated with the package. In \hyperref[sec:fit]{Section 3}, the bulk of the functions are presented, including the model building and prediction functions; some visualization tools are also showcased with accompanying examples. In \hyperref[sec:cv]{Section 4}, helper functions such as those for cross-validation and tuning are detailed. Lastly, \hyperref[sec:conc]{Section 5} includes concluding thoughts and scheduled updates for the package.

\section[Theoretical Background]{Functional neural networks (FNN)} \label{sec:bg}

The conventional neural network is a contiguous combination of hidden layers, each of which contains some number of neurons. Let $n_{u}$ denote the number of neurons in the $u$-th hidden layer. We can define the first hidden layer (denoted by the superscript) as: $\boldsymbol{v}^{(1)} = g\left(\boldsymbol{W}^{(1)}\boldsymbol{x} + \boldsymbol{b}^{(1)}\right),$ where $\boldsymbol{x}$ is a vector of $J$ covariates, $\boldsymbol{W}^{(1)}$ is an $n_1 \times J$ weight matrix, $\boldsymbol{b}^{(1)}$ is the bias (or the intercept), and $g:\mathbb{R}^{n_1} \rightarrow \mathbb{R}^{n_1} $ is some activation function that transforms the resulting linear combination \citep{ESL}. The vector $\boldsymbol{v}^{(1)}$ is $n_{1}$-dimensional and becomes the input to the next layer i.e., it takes the place of the $\boldsymbol{x}$ in the proceeding layer.

Observe that the classic neural network framework only permits the vectors (of the $N$ observations) $\{\boldsymbol{x}_{1}, \boldsymbol{x}_{2}, ...,$ $\boldsymbol{x}_{N}\}$ to contain elements that are finite scalar values. We now consider the case when $\boldsymbol{x}_{\ell}: \ell \in \{1, ...,N\}$ can be a combination of functional and scalar inputs. In this scenario, the input space takes the following form:
\begin{align*}
    \text{input}_\ell = \{x_{1}(t), x_{2}(t), ..., x_{K}(t), z_{1}, z_{2}, ..., z_{J}\},
\end{align*}
where $K$ and $J$ are the number of functional and scalar covariates, respectively. This set is unique for each of the $N$ observations. With this generalization, the $i$-th neuron of the first hidden layer corresponding to each of the $\ell$ observations takes the form \citep{thind2020deep}:
\begin{align*}
    v_i^{(1)} = g\left(\sum_{k = 1}^{K}\int_{\mathcal T_{k}} \beta_{ik}(t)x_{k}(t)dt + \sum_{j = 1}^{J}w_{ij}^{(1)}z_{j} + b_i^{(1)}\right),
\end{align*}
where $\beta(t)$ is the weight on the functional covariate; this is of a functional form because each of the $k$ functional inputs must be weighted at every point along their domains, $\mathcal{T}_{k}$. In order to estimate $\beta(t)$, it is rewritten as a linear combination of basis functions:
\begin{align}
\label{eq1and2}
    v_i^{(1)} &= g\left(\sum_{k = 1}^{K}\int_{\mathcal T_{k}} \sum_{m = 1}^{M_k}c_{ikm}\phi_{ikm}(t)x_{k}(t)dt + \sum_{j = 1}^{J}w^{(1)}_{ij}z_{j} + b^{(1)}_i\right)\\
    &= g\left(\sum_{k = 1}^{K}\sum_{m = 1}^{M_k}c_{ikm}\int_{\mathcal T_{k}} \phi_{ikm}(t)x_{k}(t)dt + \sum_{j = 1}^{J}w^{(1)}_{ij}z_{j} + b^{(1)}_i\right).
\end{align}
In \hyperref[eq1and2]{Equation (1)}, $\beta_{ik}(t)$ is approximated by $\sum_{m = 1}^{M_k}c_{ikm}\phi_{ikm}(t)$ where $\boldsymbol{\phi_{ik}(t)}$ is the set of basis functions and $\boldsymbol{c_{ik}}$ is the set of basis coefficients to be estimated by the functional neural network; in \hyperref[eq1and2]{Equation (2)}, the integral and sum are swapped and the term $\int_{\mathcal T_{k}} \phi_{ikm}(t)x_{k}(t)dt$ can be approximated using any of the usual approximation methods such as Simpson's rule \citep{Suli03}. Note that $x_{k}(t)$ is defined by its own set of basis functions.

A by-product of this method is the set of functional weights, $\beta(t)$, as represented by their basis expansions. These weights differ in comparison to the usual weights (found in conventional neural networks) in that they can be easily visualized over the corresponding continuum. This visualization illuminates the underlying relationship between the functional covariate and the scalar response. Moreover, passing in information in this functional form preserves the autocorrelation structure associated with the data. Due to the random initializations in the conventional neural network, these properties of the data become difficult to maintain if passed in discretely. In the canonical cases where we have more than one neuron, the average of the functional weights $\hat{\beta}_{k}(t) = \sum_{i=1}^{n_1}\hat{\beta}_{ik}(t)/{n_1}$, is used as the output.

This formulation of the neural network was shown \citep{thind2020deep} to be a universal approximator just as conventional neural networks are \citep{cybenko1989approximation}. While the details are omitted, the gist of the proof was to show that this form of the neuron, once simplified, held the same properties as the usual neural network.

\section[Fitting, Prediction, and Visualization]{Fitting, prediction, and visualization} \label{sec:fit}

\subsection{Data description}

Throughout this paper, we will focus on three main data sets: the \code{tecator} data set, the \code{gasoline} data set and the \code{gait} data set. The classic \code{tecator} data from the \pkg{fda.usc} package provides measurements of the near infrared absorbance spectrum and the moisture (water), fat and protein contents of 215 meat samples. Samples composed of different fat, water and protein contents may behave inconsistently in incident radiation absorption, which can be measured by absorbance spectroscopy. The data were recorded on a tecator infratec food and feed analyzer with wavelength range from 850 nm to 1050 nm using the near infrared transmission (NIT) principle. This data set consists of a list named \code{absorp.fdata} and a simple data frame labelled with \code{y}. The list \code{absorp.fdata} includes the matrix \code{data} summarizing a 100 channel spectrum of absorbance of each meat sample, where absorbance is $-\log_{10}$ of the transmittance determined by the spectrometer, and the vector \code{argvals} which contains the value of the 100 discretized channels from 850 nm to 1050 nm. The data frame \code{y} lays out the contents of the fat, water and protein of the 215 meat samples that are determined by analytic chemistry and measured in percent. This data has been applied with the goal of classifying the fat content levels of the meat samples with the functional covariate of the near infrared absorbance spectrum and the scalar covariate of the water content.

Secondly, we consider the \code{gasoline} data set sourced from the \pkg{refund} package \citep{refund}. This data set contains the near infrared reflectance (NIR) spectra and the specified octane numbers of 60 gasoline samples. For each gasoline sample,  401 NIR spectra were measured at different wavelengths, from 900 nm to 1700 nm in 2-nm intervals, and recorded using diffuse reflectance as $\log(1/\text{reflectance})$. As a simple data frame, the \code{gasoline} data consists of a vector containing the octane levels for each of the 60 collected samples, and a $60 \times 401$ matrix regarding the details of NIR measurements.  This data set has been used in examples with the aim of determining the octane number, a scalar value, of gasoline based on the functional predictor of the NIR measurements. Studies in past years \citep{gasoline_prediction, gasoline_determination} have supported the concept that a relationship exists between the chemical structure and the fuel-performance property of octane, while NIR spectroscopy is a popular method for simultaneous chemical analysis. Therefore, it has been implied that the properties of gasoline with different octane levels can be reflected by different NIR spectral features.

Finally, the examples given in this paper also make use of the \code{gait} data set obtained from the \pkg{fda} package. This data catalogs the hip and knee angles measured in degree through a twenty-point gait cycle for 39 boys. It is in the form of an array of dimension $20 \times 39 \times 2$, indicating the 20 evenly-spaced standardized gait times (from 0.025 to 0.975), the 39 subjects of observation, and the two gait variables Hip Angle and Knee Angle, respectively. It is unsurprisingly apparent that hip rotation has a regular association with knee movement in a normal walk. Therefore, in our example, we consider using hip movement as a functional covariate to make a prediction for the functional response knee angle.

% \begin{leftbar}
%  The daily( format) data set from \pkg{fda} package \citep{Ramsay05, ramsay2009} has been widely used in researches related to functional data analysis. This data set describes the daily temperature and precipitation at 35 different cities in Canada averaged over 1960 to 1994. The daily data set is a list with three components, including a vector place providing the names of the 35 Canadian cities where the weather data were collected, a $365 \times 35$ matrix tempav( format) giving the average temperature in degrees Celsius for each continuous day of the year rounded to $0.1^\circ$C, and a $365 \times 35$ matrix precav( format) offering the average daily rainfall for each day of the year rounded to 0.1 mm.
% \end{leftbar}

\subsection{Pre-processing}

The \pkg{FuncNN} package allows two types of inputs of data; data can be passed in raw or pre-processed. As with most functional data analysis coding, pre-processing is required to approximate the raw data $x(t)$, using basis expansions. However, this can be tedious and ultimately, a barrier for anyone who is not familiar with the required procedures. The \pkg{FuncNN} package provides a remedy by letting novice users input their data without the heavy lifting of converting it into its functional form while also granting the opportunity for further customization for users who require a more specific or specialized pre-processing. In the case where the data is passed in raw, the \pkg{FuncNN} package will do a simple pre-processing using a Fourier basis expansion with 31 terms. While this is an average-case compromise that minimizes initial effort, the alternative approach of pre-processing allows for greater flexibility in the development of $x(t)$.

Since the functional observations can be passed in two different ways, they can be one of two object types. In the raw case, the functional covariates are passed in as a $K$-dimensional list where each element of the list is an $N \times p$ \fct{data.frame} containing $p$ measurements over the domain for each of the $N$ subjects. 
\begin{CodeChunk}
\begin{CodeInput}
R>  func_cov_list = list(func_cov_1, func_cov_2, ..., func_cov_K)
\end{CodeInput}
\end{CodeChunk}
The scalar covariates can be passed in similar to most predictive functions i.e., as an $N \times J$ matrix containing information on the $J$ scalar variables for each of the $N$ observations.

If pre-processing is done beforehand, then the object being passed in must be a tensor. In \proglang{R}, the tensor-type object exists as an \fct{array}. The dimensionality of the tensor will be $M_{k} \times N \times K$ where each row corresponds to the estimated coefficient when the basis expansion was made, each column is an individual observation, and each ``slice'' corresponds to one of the $K$ functional covariates. Note that scaling is handled internally by the \fct{fnn.fit} function. The response can be passed in as a vector of length $N$.

 While the focus here is on scalar responses, the \pkg{FuncNN} package can handle a version of functional responses; this version entails a basis expansion of the response curves from which the basis coefficients are extracted and passed in as a matrix into the model function. The model will then attempt to predict these coefficients. More details are provided later in the article.

\subsection{Classification}

\begin{figure}[h]
  \centering
  \includegraphics[height = 14em]{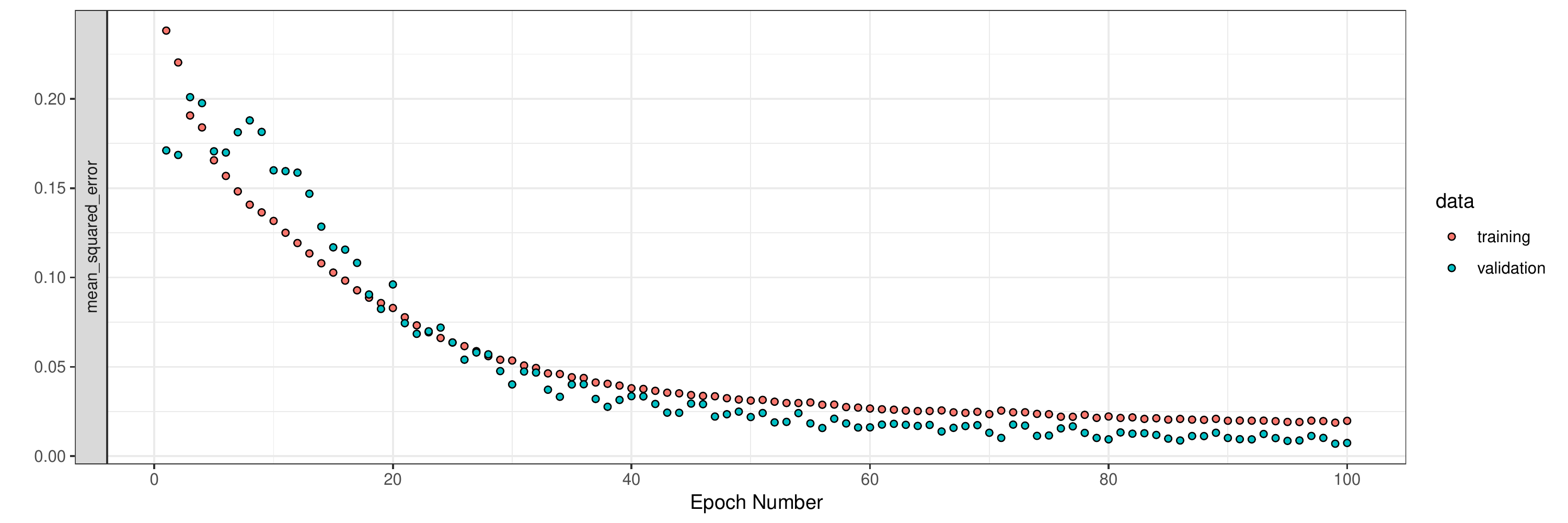}
  \label{fig:ClassTec}
  \caption{The iterative error plot over the training process as measured by mean squared error for the \code{tecator} data set.}
\end{figure}

\begin{figure}[h]
  \centering
  \includegraphics[height = 12em]{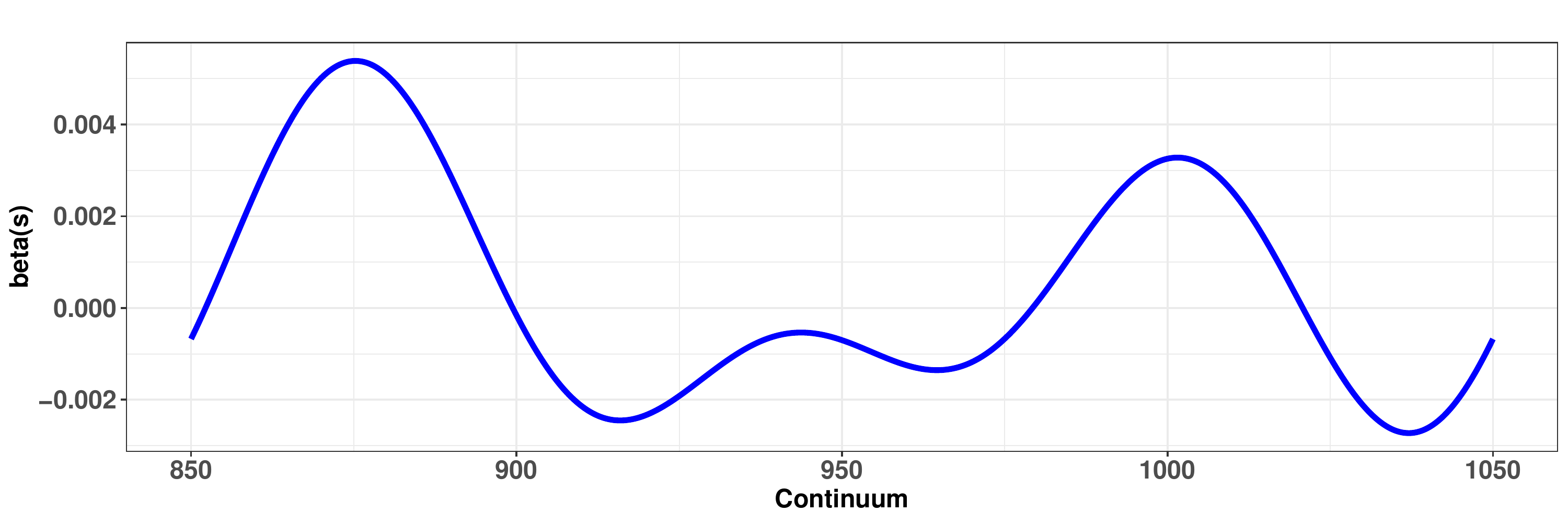}
  \label{fig:fnc_tec}
  \caption{The functional weight estimated through the functional neural network for the \code{tecator} data set. We observe two spikes. We also note that the magnitude of the effect depends on the scale of both the functional data and the functional weight.}
\end{figure}

In this example, we will focus on the \code{tecator} data set. The response will be the classification of the fat content of the meat samples as ``high'' or ``low'', denoted by 1 and 0, respectively. The threshold is 25 percent i.e., if the fat content of a sample is greater than 25, then we designate the response associated with that sample as 1:
\begin{CodeChunk}
\begin{CodeInput}
R> tecator_resp = as.factor(ifelse(tecator$y$Fat > 25, 1, 0))
\end{CodeInput}
\end{CodeChunk}
We use the water content of the meat samples as a scalar covariate; as alluded to earlier, this will be stored in a \fct{data.frame} object. 
\begin{CodeChunk}
\begin{CodeInput}
R> tecator_scalar = data.frame(water = tecator$y$Water)
\end{CodeInput}
\end{CodeChunk}
Since this is a prediction problem, we split the data into a training and test set; a random sample of 75\% of the data will be used to build the model and evaluation is performed on the rest. We will let the modelling function handle the pre-processing for this example and therefore, the object type to be passed in must be a list which can be generated as follows:
\begin{CodeChunk}
\begin{CodeInput}
R> func_covs_train = list(train_x)
R> func_covs_test = list(test_x)
\end{CodeInput}
\end{CodeChunk}
where \texttt{func\_covs\_train} and \texttt{func\_covs\_test} are the functional covariate matrices for the training and test sets, respectively (in this example, there is just one functional covariate). Finally, we can fit the model:
\begin{CodeChunk}
\begin{CodeInput}
R> fit_class = fnn.fit(resp = train_y,
+                      func_cov = func_covs_train,
+                      scalar_cov = scalar_train,
+                      domain_range = list(c(850, 1050)),
+                      raw_data = T)
\end{CodeInput}
\begin{scriptsize}
\begin{CodeOutput}
[1] "Evaluating Integrals:"
  |++++++++++++++++++++++++++++++++++++++++++++++++++| 100% elapsed=06s  
Model
_______________________________________________________________________________________________________
Layer (type)                                  Output Shape                             Param #         
=======================================================================================================
dense (Dense)                                 (None, 64)                               576             
_______________________________________________________________________________________________________
dense_1 (Dense)                               (None, 64)                               4160            
_______________________________________________________________________________________________________
dense_2 (Dense)                               (None, 2)                                130             
=======================================================================================================
Total params: 4,866
Trainable params: 4,866
Non-trainable params: 0
_______________________________________________________________________________________________________

\end{CodeOutput}
\end{scriptsize}
\end{CodeChunk}
The first item printed refers to the integral approximations that take place as shown in \hyperref[sec:bg]{Section 2}; a progress bar is printed indicating the proportion of work completed in this phase. The training error plot for this model is in \hyperref[fig:ClassTec]{Figure 1} -- note that this output is a \fct{ggplot} \citep{ggplot2} object; this is the same type of object you would extract from the \pkg{keras} package. As with any Keras model, you can observe the total number of parameters of the model in the display (in this case, 4886); this printing of information can be muted by setting \texttt{print\_info} to \code{FALSE}. There are a number of hyperparameters that will be discussed later in the article; in this simple example, we stick to the defaults. The only additional specifications required in the model are to indicate that raw data is being passed in (the default is set to \code{FALSE}) and the range of the domain. This simple syntax removes the steep learning curve that comes with functional methods in some of the other packages discussed earlier. The output contains a Keras model object as well the generated functional observations (if the raw data was passed in), parameter information, and error rates over the training iterations. Additional information about the output object can be found in the \proglang{R} documentation for the \pkg{FuncNN} package.
We now have a fitted model to make predictions with. As with most predictive functions, when one uses \fct{fnn.predict}, the fitted model is passed in along with the ``new data'' e.g., the \texttt{func\_covs\_test} object from before (along with the scalar counterpart).
\begin{CodeChunk}
\begin{CodeInput}
R> predict_class = fnn.predict(fit_class,
+                              func_cov = func_covs_test,
+                              scalar_cov = scalar_test,
+                              domain_range = list(c(850, 1050)),
+                              raw_data = T)
\end{CodeInput}
\end{CodeChunk}
The syntax is similar again. The function will automatically recognize that this is a classification problem and output class probabilities (the sum of each row will be 1). We can round these probabilities to get classifications and plot the confusion matrix (using the \pkg{caret} package \citep{caret}); this is showcased in the following code chunk:
\begin{CodeChunk}
\begin{CodeInput}
R> caret::confusionMatrix(as.factor(rounded_preds), as.factor(test_y))
\end{CodeInput}
\begin{CodeOutput}
Confusion Matrix and Statistics

          Reference
Prediction  0  1
         0 40  0
         1  1 13
                                          
               Accuracy : 0.9815          
                 95% CI : (0.9011, 0.9995)
    No Information Rate : 0.7593          
    P-Value [Acc > NIR] : 6.299e-06       
                                          
                  Kappa : 0.9506          
                                          
 Mcnemar's Test P-Value : 1               
                                          
            Sensitivity : 0.9756          
            Specificity : 1.0000          
         Pos Pred Value : 1.0000          
         Neg Pred Value : 0.9286
\end{CodeOutput}
\end{CodeChunk}
We can also visualize the relationship between the functional covariates and the scalar response by way of the functional weights using the \fct{fnn.fnc} function. The syntax is fairly straight-forward:
\begin{CodeChunk}
\begin{CodeInput}
R> fnc_tec = fnn.fnc(fit_class, domain_range = list(c(850, 1050))) 
\end{CodeInput}
\end{CodeChunk}
As a point of interest, the \textit{fnc} part of the \fct{fnn.fnc} function stands for functional neural coefficient; this is legacy code naming convention but as a reference, anytime this is mentioned, it refers to the functional weights of the model. For this function, we only need to pass in the model along with the domain. The output will be the effect of the functional weight on the response over the continuum as estimated by the functional neural network; these can be made sense of at times, however, contextual information is required. We do hazard that because of the plethora of other parameters in the network, it can be difficult to interpret these parameters -- these interpretations should be considered carefully on a case-by-case basis. In this example, we observe that there seems to be two primary spikes over the continuum. The resulting functional weight from this model is given in \hyperref[fig:fnc_tec]{Figure 2}.

\subsection{Regression}

We now consider an example of a regression problem; we will investigate the relationship between the near-infrared reflectance spectra (functional covariate) and and octane values (scalar response) of 60 gasoline observations in the \code{gasoline} data set. In this example, we will process the raw data beforehand and then pass it into \fct{fnn.fit}; we will also use multiple functional covariates by taking the derivatives of the functional data estimated using the raw data. In order to generate the functional observations, we use 9 Fourier basis functions. The \fct{Data2fd} function from the \pkg{fda.usc} package will allow us to create these functions with the following code:
\begin{CodeChunk}
\begin{CodeInput}
R> nbasis = 9
R> spline_basis = create.fourier.basis(c(900, 1700), nbasis)
R> gasoline_fd = Data2fd(timepts, t(gasoline[,-1]), spline_basis)
R> gasoline_deriv1 = deriv.fd(gasoline_fd)
R> gasoline_deriv2 = deriv.fd(gasoline_deriv1)
\end{CodeInput}
\end{CodeChunk}
%R> # convert the functional predictor into a fda object
%
where \texttt{gasoline\_fd} is the functional data object converted from the functional predictor. As alluded to earlier, when the data is pre-processed manually out of the function, we have to pass it in as a tensor object i.e., \texttt{arrays} in \proglang{R}. We can develop the array with the following code:
\begin{CodeChunk}
\begin{CodeInput}
R> gasoline_data = array(dim = c(nbasis, 60, 3))
R> gasoline_data[,,1] = gasoline_fd$coefs
R> gasoline_data[,,2] = gasoline_deriv1$coefs
R> gasoline_data[,,3] = gasoline_deriv2$coefs
\end{CodeInput}
\end{CodeChunk}

We now fit the model using a functional weight defined with five basis functions. In this initial release, we allow users to use Fourier and B-spline basis functions for the functional weights. Note that we specified the learning rate and the number of epochs in this model as well, along with the number of layers -- these are just some of the numerous hyperparameters that can be adjusted in this function. 
\begin{CodeChunk}
\begin{CodeInput}
R> gasoline_example <- fnn.fit(resp = train_y, 
+                              func_cov = gasoline_data_train, 
+                              scalar_cov = NULL,
+                              basis_choice = c("bspline"), 
+                              num_basis = c(5),
+                              hidden_layers = 2,
+                              neurons_per_layer = c(64, 64),
+                              activations_in_layers = c("relu", "linear"),
+                              domain_range = list(c(900, 1700)),
+                              epochs = 300,
+                              learn_rate = 0.0001,
+                              early_stopping = T)
\end{CodeInput}
\begin{scriptsize}
\begin{CodeOutput}
[1] "Warning: You only specified basis information for one functional covariate -- 
it will be repeated for all functional covariates"
[1] "Evaluating Integrals:"
  |++++++++++++++++++++++++++++++++++++++++++++++++++| 100% elapsed=01s  
  |++++++++++++++++++++++++++++++++++++++++++++++++++| 100% elapsed=01s  
  |++++++++++++++++++++++++++++++++++++++++++++++++++| 100% elapsed=01s  
Model
_______________________________________________________________________________________________________
Layer (type)                                  Output Shape                             Param #         
=======================================================================================================
dense (Dense)                                 (None, 64)                               1024            
_______________________________________________________________________________________________________
dense_1 (Dense)                               (None, 64)                               4160            
_______________________________________________________________________________________________________
dense_2 (Dense)                               (None, 1)                                65              
=======================================================================================================
Total params: 5,249
Trainable params: 5,249
Non-trainable params: 0
_______________________________________________________________________________________________________
\end{CodeOutput}
\end{scriptsize}
\end{CodeChunk}
Observe the warning present in the output -- the \pkg{FuncNN} package has many stop checks built to provide clear information to the user regarding errors and warnings. In this case, we had 3 functional covariates but only specified information for one functional weight - i.e., we specified the functional weight to be a B-spline expansion of five terms.  To make up for the lack of information, \fct{fnn.fit} will just repeat this same setup for all of the inputted functional covariates. 

\begin{figure}[h]
  \centering
  \includegraphics[height = 14em]{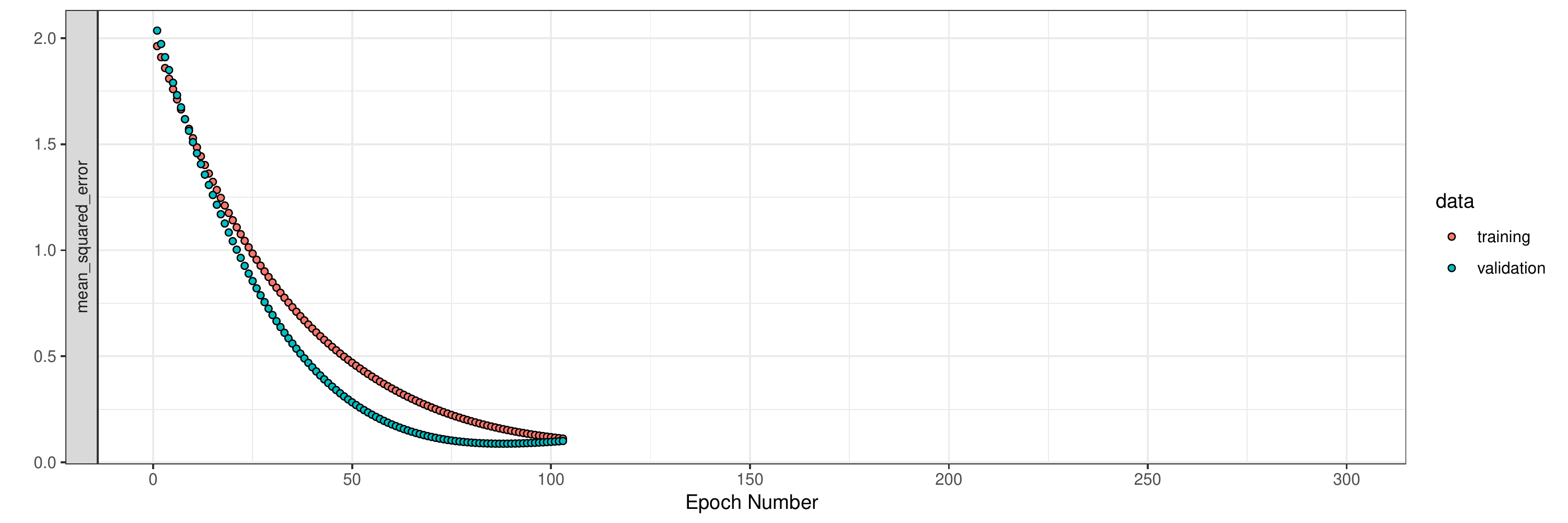}
  \label{fig:RegGas}
  \caption{The iterative error plot over the training process as measured by mean squared error for the \code{gasoline} data set.}
\end{figure}

Again, we provide the Keras model information. We observe that there were 5249 parameters being trained during the model building process. The error plot is given in \hyperref[fig:RegGas]{Figure 3} -- observe that the \texttt{early\_stopping} parameter was triggered in this case so the plot does not extend to the specified 300 epochs.  More information on this parameter is provided later in the article.
Predictions are handled using the same function, \fct{fnn.predict}; again, the only difference will be that the output will be a vector of values as opposed to a matrix containing class probabilities. 

\subsection{Functional responses}

To demonstrate functional responses, we will predict knee movement using hip movement with the information presented in the \code{gait} data set. While we present a straightforward approach to functional response prediction, we will update the package with a more refined and sophisticated approach as we continue to develop such methodology.  The approach used in the following example boils down to predicting the coefficients of the functional data object corresponding to the response. To begin, we read in the data and generate our functional observations:
\begin{CodeChunk}
\begin{CodeInput}
R> hipbasis13 = create.fourier.basis(c(0,1), 13)
R> kneebasis11 = create.fourier.basis(c(0,1), 11)
R> timepts = as.numeric(rownames(gait))
R> hip_fd = Data2fd(timepts, gait[,,1], hipbasis13)
R> knee_fd = Data2fd(timepts, gait[,,2], kneebasis11)
\end{CodeInput}
\end{CodeChunk}
In this example, we use 13 and 11 basis functions to generate the hip and knee observations, respectively. The pre-processing is done in a manner similar to that of the regression case above. A notable point is that the response will now be a $N \times M_{\text{resp}}$ matrix object where $N$ is the number of functional observations and $M_{\text{resp}}$ is the number of basis functions used to define these observations -- the values in the matrix correspond to the estimated basis coefficients (in this example, we have 11 per observation). 

We now fit the model using four layers containing 64 neurons each with \textit{relu} \citep{hahnloser2000digital} activation function. The learning rate was selected after a small grid search. 
\begin{CodeChunk}
\begin{small}
\begin{CodeInput}
R> gait_fit <- fnn.fit(resp = resp_train,
+                      func_cov = data_train,
+                      hidden_layers = 3,
+                      activations_in_layers = c("relu", "relu", "relu"),
+                      neurons_per_layer = c(128, 128, 32),
+                      epochs = 300,
+                      learn_rate = 0.0007)
\end{CodeInput}
\end{small}
\begin{scriptsize}
\begin{CodeOutput}
[1] "Evaluating Integrals:"
  |++++++++++++++++++++++++++++++++++++++++++++++++++| 100% elapsed=02s  
Model
_______________________________________________________________________________________________________
Layer (type)                                  Output Shape                             Param #         
=======================================================================================================
dense (Dense)                                 (None, 128)                              1024            
_______________________________________________________________________________________________________
dense_1 (Dense)                               (None, 128)                              16512           
_______________________________________________________________________________________________________
dense_2 (Dense)                               (None, 32)                               4128            
_______________________________________________________________________________________________________
dense_3 (Dense)                               (None, 11)                               363             
=======================================================================================================
Total params: 22,027
Trainable params: 22,027
Non-trainable params: 0
_______________________________________________________________________________________________________
\end{CodeOutput}
\end{scriptsize}
\end{CodeChunk}
\begin{figure}[h]
  \centering
  \includegraphics[height = 14em]{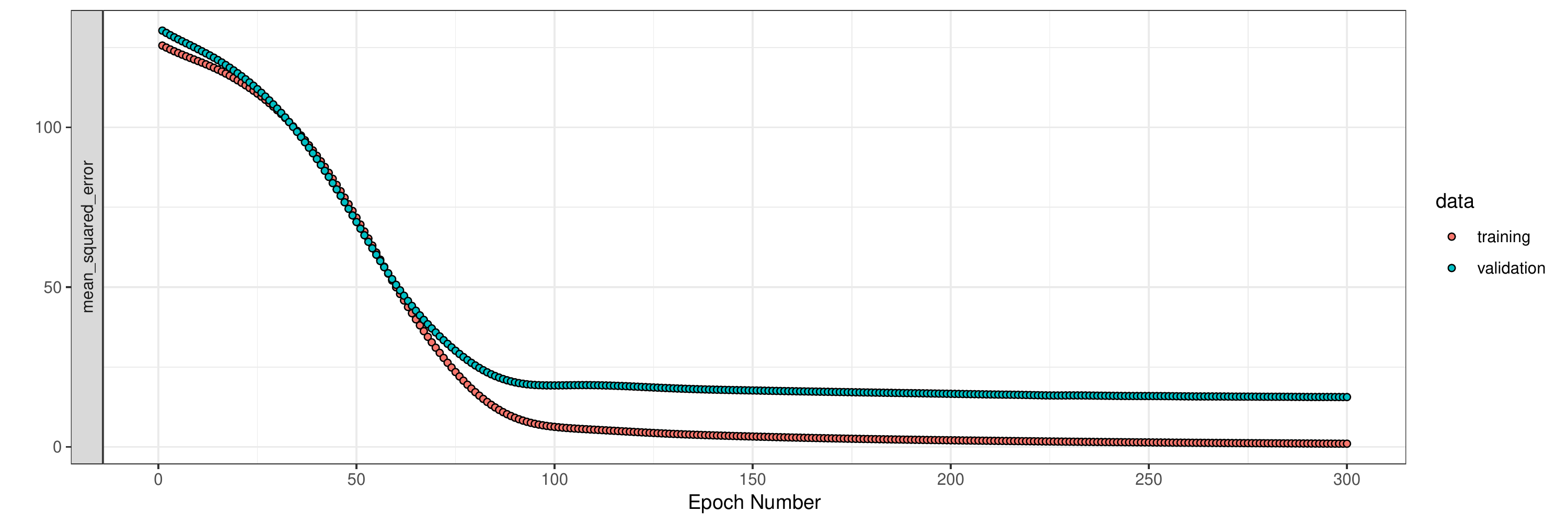}
  \label{fig:funcRespError}
  \caption{The iterative error plot over the training process as measured by mean squared error for the \code{gait} data set.}
\end{figure}
The error plot corresponding to this model is given in \hyperref[fig:funcRespError]{Figure 4}.  This model was, thus far, the most parameter heavy of the examples with a total of 22,027. Predictions are made in a similar manner as before; the output object will be a matrix with a number of columns equal to that of $M_{\text{resp}}$ and a row dimensionality the same as the number of functional observations in the test set.
\begin{CodeChunk}
\begin{CodeInput}
R> predictions = fnn.predict(model = gait_fit, 
+                            func_cov = data_test)
\end{CodeInput}
\end{CodeChunk}
 To visualize the results, the \fct{fnn.plot} function can be used which will output the evaluations of the functional response prediction over the domain. The syntax consists of specifying the prediction object, the domain, a step size which determines how finely evaluations are made, and the type of basis function to be used for the plot.
\begin{CodeChunk}
\begin{CodeInput}
R> gait_func_pred = fnn.plot(predictions, 
+                            domain_range = c(0, 1), 
+                            step_size = 0.05, 
+                            Basis_Type = "fourier")
\end{CodeInput}
\end{CodeChunk}
Using these predictions, we can compare our predicted curve to the true functional response; \hyperref[fig:funcResp]{Figure 5} shows one such comparison. 

\begin{figure}[h]
  \centering
  \includegraphics[height = 12em]{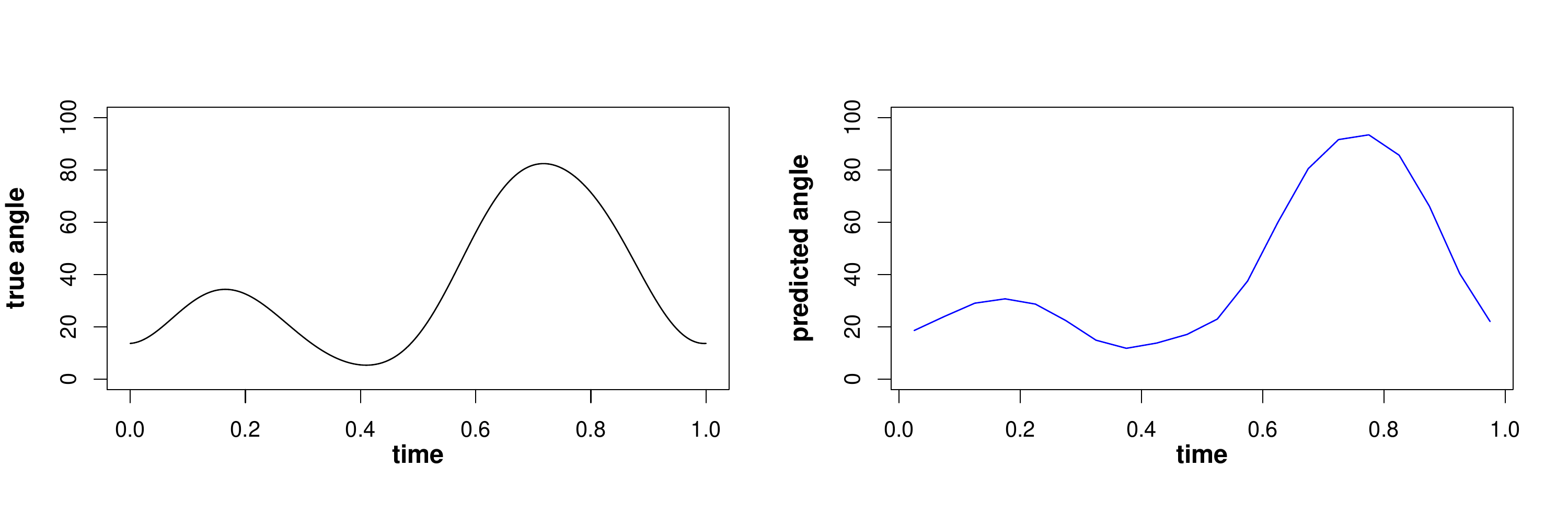}
  \label{fig:funcResp}
  \caption{A comparison of a knee angle prediction $\hat{y}(t)$ over a 1 second interval (functional response) with the observed movement, $y(t)$.}
\end{figure}
%8x4.5 plots

\section[Tuning and CV]{Model improvement tools} \label{sec:cv}

\subsection{Hyperparameters}

Thus far, we have generated models largely using the default settings of the \pkg{FuncNN} package; however, there is ample opportunity to adjust these models so that they are best suited for a particular context. These adjustments come by way of toying with the hyperparameter options available to us in the modelling process.
\begin{center}
\setlength{\tabcolsep}{15pt}
\setlength{\columnseprule}{0.2pt}
\scalebox{0.65}{
\label{table:one}
 \begin{tabular}{||c | c | c ||} 
 \hline
 \texttt{Parameter} & \texttt{Type} & \texttt{Details} \\ [1ex] 
 \hline\hline
  $\beta(t)$ & Estimated & Coefficient function found by the FNN. \\ [1ex] 
  \hline\hline
  $w$ & Estimated & The scalar covariate weights. \\ [1ex] 
 \hline\hline
  $b$ & Estimated & The intercept in each neuron. \\ [1ex] 
 \hline\hline
  Number of lyers & Hyperparameter & The depth of the FNN. \\ [1ex] 
 \hline\hline
  Neurons per layer & Hyperparameter  & Number of neurons in each layer of the FNN. \\ [1ex] 
 \hline\hline
  Learn rate & Hyperparameter & The learning rate of the FNN. \\ [1ex] 
 \hline\hline
  Decay rate & Hyperparameter & A weight on the learning process across training iterations for the FNN. \\ [1ex] 
  \hline\hline
  Validation split & Hyperparameter & The split of training set during learning process. \\ [1ex] 
  \hline\hline
  Functional weight basis & Hyperparameter & The size of $M$ for the estimation of $\beta(t)$. \\ [1ex] 
  \hline\hline
  Training iterations & Hyperparameter & The number of learning iterations. \\ [1ex] 
  \hline\hline
  Batch size & Hyperparameter & Subset of data per pass of the FNN. \\ [1ex] 
  \hline\hline
  Activations & Hyperparameter & The choice of $g(\cdot)$ for each layer. \\ [1ex] 
  \hline\hline
  Early stopping & Hyperparameter & Stops the model building process if no improvement in error. \\ [1ex]
  \hline\hline
  Dropout & Hyperparameter & Randomly drops some specified percentage of neurons from one layer to the next. \\ [1ex]
  \hline\hline
\end{tabular}}
\captionof{table}{A list of the parameters in the network.}
\end{center}

Some parameters, such as the ones seen explicitly in \hyperref[eq1and2]{Equation 1}, are estimated through gradient descent - i.e., the "learning" process of the model results in the final values. Others, such as the number of hidden layers and the neurons within each of these layers, are adjusted for by the user. For example, as demonstrated earlier in the article, the \texttt{early\_stopping} option halts the learning process when significant decreases in error have not been observed for some number of training iterations -- this number, known as the patience parameter, is also a hyperparameter. Another example is that of the learning rate, $\gamma$, and the decay rate, which both govern how quickly we move in the direction of the gradient. The complete list of hyperparameters along with descriptions is provided in \hyperref[table:one]{Table 1}.

\subsection{Cross-validation}

A first step in choosing parameter values in any modelling process is to find a reliable measuring stick to assess how a model performs. One approach to solving this problem is cross-validation. The general definition for a k-fold cross-validation error is \vspace{-.2cm} $$\text{MSPE} = \sum_{k=1}^K \sum_{l \in S_k}\left(\hat{y}_{l}^{(-k)} - y_{l}\right)^2/\left(|S_k|K\right) \vspace{-.2cm}$$  where $\{S_k\}_{k=1}^K$ is a partition of the data set, $|S_k|$ denotes cardinality, and $\hat{y}_{l}^{(-k)}, l \in S_k,$ is the predicted value for $y_l$ obtained by training the functional neural network using data points with indices not contained in $S_k$.  Choosing to measure error in this way ensures that our measurement is computed using data that was not used to train the model, reducing the bias relative to other estimations of test error.

This paradigm is at the core of \fct{fnn.cv}, the package's cross-validation function. Much of the syntax overlaps with the \fct{fnn.fit} function. We demonstrate the cross-validation function using the \code{gasoline} data set. Instead of splitting the data into test and training sets, we can pass in the entire data set -- the cross-validation function will take care of this internally. The \texttt{nfolds} option determines the number of subsets in our partition of the full data set -- this is the only syntax difference between the \fct{fnn.fit} and \fct{fnn.cv} functions.
\begin{CodeChunk}
\begin{CodeInput}
R> gasoline_cv <- fnn.cv(nfolds = 10,
+                        resp = octane, 
+                        func_cov = gasoline_data, 
+                        scalar_cov = NULL,
+                        basis_choice = c("bspline", "bspline", "bspline"), 
+                        num_basis = c(5, 5, 5),
+                        hidden_layers = 2,
+                        neurons_per_layer = c(64, 64),
+                        activations_in_layers = c("relu", "linear"),
+                        domain_range = list(c(900, 1700), 
+                                           c(900, 1700), 
+                                           c(900, 1700)),
+                        epochs = 300,
+                        learn_rate = 0.0001,
+                        early_stopping = T)
\end{CodeInput}
\end{CodeChunk}
While the function is running, the number of completed folds will be displayed and updated, for example:
\begin{CodeChunk}
\begin{CodeOutput}
[1] "Folds Done: 1"
[1] "Folds Done: 2"
...
\end{CodeOutput}
\end{CodeChunk}
The output will contain the overall cross-validated error as well as the errors corresponding to each fold. The indices for each fold split will also be available for users if they choose to reproduce the function results manually. In this example, a 10-fold cross-validation was done and the MSPE was 0.0735. We look to improve on this in the next section.

\subsection{Tuning}

Having honed in on a more robust measure of error, we can begin to outline the tuning approach, which consists of a classic grid search to develop testable combinations of hyperparameters. Algorithm 1 below details the tuning process used in \pkg{FuncNN}.
% Firstly, in order to trigger this tuning process, users are required to manually define a subset of all possible values for each of the hyperparameters of interests. With the specified grid, many versions of the designed functional neural network will be constructed with all possible combinations of hyperparameters. Afterwards, with a given number of folds, the cross-validation process described earlier will be applied to measure the MSPE, a typical evaluation metric to guide the optimization algorithm, for training each of the functional neural networks. The tuning approach will finally select and return the set of hyperparameters with the smallest MSPE, which can be referred to as the best performance. 
%
\IncMargin{1.5em}
\begin{algorithm}
\SetAlgoNoLine
\DontPrintSemicolon
\newcommand{\hrulealg}[0]{\vspace{2mm} \hrule \vspace{1mm}}
\SetKwBlock{Loop}{For}{End}
\linespread{1.15}\selectfont

\KwIn{A subset of all possible values for each of the hyperparameters of interest, including: number of hidden layers, number of neurons, number of training iterations, percentage of split, patience, learning rate, number of basis functions, and activation function}
\KwOut {The optimal set of hyperparameters ($\eta$) over all combinations of the elements in \textbf{Input}} 
\hrulealg
\nl Input the grid, along with the functional and scalar observations, and set the non-tuning hyperparameters, e.g., type of basis functions, number of folds ($k$) for cross-validation, rate of decay, etc.\\
\nl Generate all possible combinations of hyperparameters, denoted $\eta_{1}, \eta_{2}, ..., \eta_{n}$, where $n$ is the total number of combinations \\
\Loop (each hyperparameter combination $\eta_{i}, i = 1, ...., n$){
\nl Construct the functional neural network with $\eta_{i}$ \\
\nl Train the designed functional neural network\\
\nl Calculate the $k$-fold cross-validation error MSPE($\eta_{i}$)
}
\nl Return $\eta = \{\eta_{k}: $ MSPE($\eta_{k}$) $ \leq $ MSPE($\eta_{i}$)$ \text{ for all } i\}$

\caption{Tuning (Hyperparameter optimization)}
\end{algorithm}
\DecMargin{1.5em}

The \pkg{FuncNN} package makes this intuitive and easy. There is no automated pre-processing for the tuning function, \fct{fnn.tune}; this decision was made due to efficiency concerns but will be remedied for in the future. The syntax is as follows:
\begin{CodeChunk}
\begin{CodeOutput}
fnn.tune(tune_list,
         resp,
         func_cov,
         scalar_cov = NULL,
         basis_choice,
         domain_range,
         batch_size = 32,
         decay_rate = 0,
         nfolds = 5,
         cores = 4,
         raw_data = F)
\end{CodeOutput}
\end{CodeChunk}
Some of these have already been presented before. The ones to note are the \texttt{tune\_list} and \texttt{cores} parameters. The latter is an efficiency parameter that allows for parallelization. This is handled internally by the dependency \pkg{future} \citep{future}. The former must be a list object that contains the grid of combinations to be tested. An example of what this list might look like is:
\begin{CodeChunk}
\begin{CodeInput}
R> tune_list_gasoline = list(num_hidden_layers = c(2),
+                            neurons = c(32, 64),
+                            epochs = c(250),
+                            val_split = c(0.2),
+                            patience = c(15),
+                            learn_rate = c(seq(0.0001, 0.001, 
+                                           length.out = 5)),
+                            num_basis = c(5, 7, 9),
+                            activation_choice = c("relu", "sigmoid"))
\end{CodeInput}
\end{CodeChunk}
This object is not processed as obviously as it may seem. We assign each of the hidden layer choices (in this case, just the one choice of 2 hidden layers) all combinations of neurons and activation functions. This is exemplified below:
\begin{CodeChunk}
\begin{CodeOutput}
   L1_Act L2_Act FW_1 FW_2 FW_3 L1_N L2_N Epochs ValSplit Patience LearnRate
1    relu   relu    5    5    5   32   32    250      0.2       15  0.000100
2    relu   relu    5    5    7   32   64    250      0.2       15  0.000550
3    relu   relu    5    5    5   32   64    250      0.2       15  0.000550
4    relu   relu    5    5    5   32   64    250      0.2       15  0.001000
5    relu   relu    5    5    5   32   64    250      0.2       15  0.000100
6    relu   relu    5    5    7   32   32    250      0.2       15  0.000325
7    relu   relu    5    5    5   32   32    250      0.2       15  0.000325
8    relu   relu    5    5    5   64   32    250      0.2       15  0.001000
9    relu   relu    5    5    5   32   32    250      0.2       15  0.000775
10   relu   relu    5    5    5   64   32    250      0.2       15  0.000100
11   relu   relu    5    5    9   32   32    250      0.2       15  0.000550
12   relu   relu    5    5    7   32   32    250      0.2       15  0.000550
13   relu   relu    5    5    5   32   32    250      0.2       15  0.000550
14   relu   relu    5    5    5   32   32    250      0.2       15  0.001000
15   relu   relu    5    5    5   64   32    250      0.2       15  0.000325
\end{CodeOutput}
\end{CodeChunk}
Observe that we are cycling through all of the possible combinations of neuron numbers, activation functions, and functional weight basis counts from the list. We did not need to specify the number of neurons for each layer separately. While this is less obvious of an approach than allowing the user to set the grid for each layer, it is much more efficient when one considers that they may want to be tuning over many layers. For example, tuning over 10 hidden layers in the traditional approach would require a list object that contains over 30 elements, whereas in this case, the length of the list remains the same no matter how many layers you are attempting to tune for.

Having set up the tuning list, we can now consider the code required to run the tuning function:
\begin{CodeChunk}
\begin{CodeInput}
R> gasoline_tuned = fnn.tune(tune_list = tune_list_gasoline,
+                            resp = octane,
+                            func_cov = gasoline_data,
+                            basis_choice = c("fourier", "fourier", 
+                                             "fourier"),
+                            domain_range = list(c(900, 1700), 
+                                                c(900, 1700), 
+                                                c(900, 1700)),
+                            nfolds = 2)
\end{CodeInput}
\end{CodeChunk}
With respect to \texttt{nfolds}, we see that the tuning comparisons will be done with a 2-fold cross-validated error. Increasing this number, as expected, will increase the computational time of this function. You may tune for a multiple number of layers at once. In that case, the grid above is reproduced for each hidden layer. The output will be a list object containing the final grid (as seen above), the cross-validated MSPE values for every combination, the best from each choice of layers, and the overall best across all layers (along with the corresponding parameters).

While the functions runs, we took advantage of the \pkg{pbapply} package \citep{pbapply} to provide users with a progress bar.
\begin{CodeChunk}
\begin{CodeOutput}
|++++++++++++++++++++++++++++++++++++++++++++++++  | 94% ~3m 49s      
\end{CodeOutput}
\end{CodeChunk}
This progress bar displays the approximate amount of time remaining until the tuning is finished; this information offers some obvious benefits to a user's quality of experience.

Our goal for this tuning (applied to the \code{gasoline} data set) was to significantly outperform the cross-validated error from the default settings presented before. After running the tuning function, the optimal parameters over the pre-defined grid were:
\begin{CodeChunk}
\begin{CodeInput}
R> gasoline_tuned$Parameters
\end{CodeInput}
\begin{CodeOutput}
$MSPE
[1] 0.002492664

$num_basis
[1] 7 9 9

$hidden_layers
[1] 2

$neurons_per_layer
[1] 32 32

$activations_in_layers
[1] "sigmoid" "relu"   

$epochs
[1] 250

$val_split
[1] 0.2

$patience_param
[1] 15

$learn_rate
[1] 0.001
\end{CodeOutput}
\end{CodeChunk}

Using these parameter choices, we can run the cross-validation function again (with ten folds for the purposes of comparison under equivalent circumstances):
\begin{CodeChunk}
\begin{CodeInput}
R> gasoline_tuned_cv <- fnn.cv(nfolds = 10,
+                              resp = octane, 
+                              func_cov = gasoline_data, 
+                              scalar_cov = NULL,
+                              basis_choice = c("bspline", "bspline", 
+                                               "bspline"), 
+                              num_basis = c(7, 9, 9),
+                              hidden_layers = 2,
+                              neurons_per_layer = c(32, 32),
+                              activations_in_layers = c("sigmoid", "relu"),
+                              domain_range = list(c(900, 1700), 
+                                                  c(900, 1700), 
+                                                  c(900, 1700)),
+                              epochs = 250,
+                              learn_rate = 0.001,
+                              early_stopping = T)
\end{CodeInput}
\end{CodeChunk}
And the resulting cross-validated MSPE was 0.0043. The results are summarized in \hyperref[table:two]{Table 2}:
\begin{center}
\setlength{\tabcolsep}{40pt}
\setlength{\columnseprule}{0.4pt}
\scalebox{0.85}{
\label{table:two}
 \begin{tabular}{||c | c | c ||} 
 \hline
 \texttt{Model} & \texttt{CV MSPE} & \texttt{Standard error} \\ [1ex] 
 \hline\hline
  Default & 0.0735 & 0.0534 \\ [1ex] 
  \hline\hline
  Tuned & 0.0043 & 0.00275 \\ [1ex] 
 \hline
\end{tabular}}
\captionof{table}{Tuning vs. default model results.}
\end{center}
We observe a much smaller standard error on our tuned model, indicating a greater level of consistency with respect to the predictions. The high variance in the default model suggests that there is some propensity for that particular model to overfit the data; it seems that behaviour is largely contained in the tuned model.

\section[Conclusions]{Conclusion and future work} \label{sec:conc}

Functional data analysis is an emerging field where new and exciting research is being conducted through a different data paradigm than the usual multivariate lens; this paper has illustrated how the functions introduced in the \pkg{FuncNN} take advantage of this paradigm and allow users to generate useful models which apply deep learning to functional data.
%one such area of research involves a marriage of functional data analysis and deep learning. While a methodology exists for such a method, there was no software that allowed users to generally apply these methods in their respective frameworks. We introduce one such software in this paper: \pkg{FuncNN}.
Since \pkg{FuncNN} package is built on top of the Keras architecture, many of the numerous parameters available to users implementing Keras models are also available for the software introduced here.

Throughout the article, we presented a number of functions. This includes core functions such as those for model fitting (\fct{fnn.fit}) and predicting (\fct{fnn.predict}) but also additional convenience functions for easier validation of analysis (e.g., \fct{fnn.cv}, \fct{fnn.tune}). These functions were highlighted with the use of examples including those in the context of regression, classification, and functional responses.

A number of updates to the package have been scheduled including the introduction of more core Keras options. We also are working on a better parallelization of the underlying processes to help with efficiency. Lastly, while we presented one way of handling functional responses in this article, there is a novel extension that is being worked on -- this approach will also be available in future versions of \pkg{FuncNN}.

\section*{Acknowledgments}

We would like to thank Matthew Reyers, Meyappan Subbaiah, and Kevin Multani for their useful discussions during the development of the package.

%% -- Bibliography -------------------------------------------------------------
%% - References need to be provided in a .bib BibTeX database.
%% - All references should be made with \cite, \citet, \citep, \citealp etc.
%%   (and never hard-coded). See the FAQ for details.
%% - JSS-specific markup (\proglang, \pkg, \code) should be used in the .bib.
%% - Titles in the .bib should be in title case.
%% - DOIs should be included where available.

\bibliography{refs}

%% -- Appendix (if any) --------------------------------------------------------
%% - After the bibliography with page break.
%% - With proper section titles and _not_ just "Appendix".

\newpage

\begin{appendix}
\label{sec:app}

\section{Computational times for all models} 

\begin{center}
\setlength{\tabcolsep}{40pt}
\setlength{\columnseprule}{0.4pt}
\scalebox{0.85}{
\label{table:three}
 \begin{tabular}{||c | c | c ||} 
 \hline
 \texttt{Section} & \texttt{Method} & \texttt{Run time}\\ [1ex] 
 \hline\hline
  3.3 - Classification & \fct{fnn.fit} & 13.0s \\ [1ex] 
 \hline\hline
  3.3 - Classification & \fct{fnn.predict} & 2.22s \\ [1ex] 
  \hline\hline
  3.3 - Classification & \fct{fnn.fnc} & 0.170s \\ [1ex] 
  \hline\hline
  3.4 - Regression & \fct{fnn.fit} & 4.22s \\ [1ex] 
 \hline\hline
  3.4 - Regression & \fct{fnn.predict} & 2.45s \\ [1ex] 
  \hline\hline
  3.4 - Regression & \fct{fnn.fnc} & 0.410s \\ [1ex] 
  \hline\hline
  3.5 - Functional responses & \fct{fnn.fit} & 5.61s \\ [1ex] 
 \hline\hline
  3.5 - Functional responses & \fct{fnn.predict} & 0.410s \\ [1ex] 
  \hline\hline
  3.3 - Functional responses & \fct{fnn.plot} & 0.230s \\ [1ex] 
  \hline\hline
  3.5 - Functional responses & \fct{fnn.fnc} & 0.270s \\ [1ex] 
  \hline\hline
  4.2 - Cross-validation & \fct{fnn.cv} & 1m12s\\ [1ex] 
  \hline\hline
  4.3 - Tuning & \fct{fnn.tune} & 8h24m\\ [1ex] 
 \hline
\end{tabular}}
\captionof{table}{Computational run times for every function used in this article from the \pkg{FuncNN} package.}
\end{center}

% \newpage
% \section{Functional Weights: Gasoline} 
% \begin{figure}[htb!]
%   \centering
%   \includegraphics[height = 12em]{fnc_gas.pdf}
%   \label{fig:fnc_gas}
%   \caption{The functional weights estimated through the functional neural network for the \code{gasoline} data set.}
% \end{figure}

% \newpage
% \section{Functional Weight: Gait} 

% \begin{figure}[h]
%   \centering
%   \includegraphics[height = 12em]{fnc_gait.pdf}
%   \label{fig:fnc_gait}
%   \caption{The functional weight estimated through the functional neural network for the \code{gait} data set.}
% \end{figure}

\newpage

% \section[Using BibTeX]{Using \textsc{Bib}{\TeX}} \label{app:bibtex}

% \begin{leftbar}
% References need to be provided in a \textsc{Bib}{\TeX} file (\code{.bib}). All
% references should be made with \verb|\cite|, \verb|\citet|, \verb|\citep|,
% \verb|\citealp| etc.\ (and never hard-coded). This commands yield different
% formats of author-year citations and allow to include additional details (e.g.,
% pages, chapters, \dots) in brackets. In case you are not familiar with these
% commands see the JSS style FAQ for details.

% Cleaning up \textsc{Bib}{\TeX} files is a somewhat tedious task -- especially
% when acquiring the entries automatically from mixed online sources. However,
% it is important that informations are complete and presented in a consistent
% style to avoid confusions. JSS requires the following format.
% \begin{itemize}
%   \item JSS-specific markup (\verb|\proglang|, \verb|\pkg|, \verb|\code|) should
%     be used in the references.
%   \item Titles should be in title case.
%   \item Journal titles should not be abbreviated and in title case.
%   \item DOIs should be included where available.
%   \item Software should be properly cited as well. For \proglang{R} packages
%     \code{citation("pkgname")} typically provides a good starting point.
% \end{itemize}
% \end{leftbar}

\end{appendix}

%% -----------------------------------------------------------------------------

\end{document}